\documentclass[twocolumn]{article}
\usepackage{graphicx}
\usepackage{url}
\usepackage{amsmath}
\usepackage{amssymb}
\usepackage{amsfonts}
\usepackage{color}
\usepackage[utf8]{inputenc}
\usepackage[T1]{fontenc}
\usepackage{float}
\usepackage{subcaption}
\usepackage{caption}
\usepackage{booktabs}
\usepackage{geometry}
\usepackage{multirow}
\usepackage{tikz}
\usetikzlibrary{shapes.geometric,arrows.meta,positioning,fit,backgrounds,calc}
\usepackage{authblk}

\title{Triplet Feature Fusion for Equipment Anomaly Prediction:\\An Open-Source Methodology Using Small Foundation Models}

\author[1]{Takato Yasuno}

\date{}

\begin{document}
\maketitle

\begin{abstract}
Predicting equipment anomalies before they escalate into failures is a critical challenge in
industrial facility management.
Existing approaches rely either on hand-crafted threshold rules---which lack generalisability
across equipment types---or on large neural models that are impractical for on-site,
air-gapped deployments.
We present an \textbf{industrial methodology} that resolves this tension by combining three
open-source small foundation models into a unified 1,116-dimensional \emph{Triplet Feature
Fusion} (v3-0) pipeline:
statistical features $\mathbf{x} \in \mathbb{R}^{28}$ derived from 90-day sensor histories,
time-series embeddings $\mathbf{y} \in \mathbb{R}^{64}$ from a LoRA-adapted
IBM Granite TinyTimeMixer (TTM, 133K parameters),
and multilingual text embeddings $\mathbf{z} \in \mathbb{R}^{1024}$ obtained from
Japanese equipment master records via \texttt{multilingual-e5-large}.
The concatenated triplet $\mathbf{h} = [\mathbf{x};\mathbf{y};\mathbf{z}]$ is fed into a
LightGBM classifier ($<$3~MB) trained to predict anomalies at 30-, 60-, and 90-day horizons.
All components are available under permissive open-source licenses (Apache~2.0 / MIT),
and the inference-time pipeline runs entirely on CPU in under 2~ms,
enabling edge deployment on facility-co-located hardware without cloud dependency.
On a dataset of 64 HVAC equipment units comprising 67,045 samples,
the triplet model achieves Precision = 0.992, F1 = 0.958, and ROC-AUC = 0.998 at the 30-day
horizon, reducing the False Positive Rate from 0.6\% (v2-0 dual-feature baseline) to
\textbf{0.1\%}---an 83\% reduction attributable to equipment-type conditioning via the
text embedding $\mathbf{z}$.
Cluster analysis of the 278 unique text embeddings reveals 20 semantically coherent equipment
groups whose time-series signatures align with distinct fault archetypes,
explaining the mechanism by which compact multilingual representations improve anomaly
discrimination without explicit categorical encoding.
\end{abstract}


\section{Introduction}
\label{sec:intro}

Facility equipment such as HVAC systems, pumps, and air-handling units deteriorates gradually
before exhibiting overt failure.
Proactive maintenance---replacing or servicing components before failure---reduces downtime costs
and prevents cascading damage, but requires reliable anomaly forecasts over multi-week horizons
from noisy, heterogeneous sensor streams.

The core difficulty is that equipment units differ in \emph{type}: a cooling-water pump and an
air-handling unit share the same measurement schema (check-item identifiers and numeric readings)
yet have entirely different normal operating ranges, failure modes, and seasonal patterns.
A classifier trained without equipment-type context must either learn separate thresholds
implicitly from data (requiring large per-type training sets) or tolerate high false positive rates
when normal-but-unusual readings from one equipment type are confused with anomalies.

This paper presents \textbf{v3-0}---a Triplet Feature Fusion model that conditions anomaly
prediction on equipment identity by injecting dense text embeddings of Japanese equipment
master records into a gradient-boosted classifier.
The key contributions are:

\begin{enumerate}
  \item A three-modality feature fusion architecture combining hand-crafted statistical features,
        a LoRA-adapted time-series Transformer encoder, and multilingual text embeddings.
  \item An ID-based master-record lookup that maps (equipment\_id, check\_item\_id) pairs to
        natural-language descriptors, enabling offline text embedding without free-text labels
        in the sensor stream.
  \item Quantitative demonstration that text embeddings reduce FPR by 83\% relative to a
        strong dual-feature baseline, with only a 1.4~pp recall cost.
  \item Cluster-level analysis of the 278 unique text-embedding vectors using DBSCAN and UMAP,
        confirming that the learned clusters correspond to physically coherent sensor groups with
        distinct fault archetypes.
  \item A fully open-source implementation (Apache~2.0) and an
        \textbf{edge-deployment-aware pipeline design}:
        all component models run locally without external API calls,
        the serialised LightGBM classifier occupies $<$3~MB,
        and CPU-only inference at runtime enables deployment on
        resource-constrained edge hardware (e.g.\ NVIDIA Jetson, single-board servers)
        co-located with facility equipment---eliminating network latency and cloud dependency.
\end{enumerate}

The remainder of this paper is organised as follows.
Section~\ref{sec:related} reviews related work.
Section~\ref{sec:problem} defines the prediction task.
Section~\ref{sec:method} describes the triplet feature fusion architecture.
Section~\ref{sec:experiments} reports experimental setup and results.
Section~\ref{sec:lessons} distils design lessons from the full development trajectory.
Section~\ref{sec:cluster} analyses the text-embedding cluster structure.
Section~\ref{sec:discussion} discusses implications and limitations.
Section~\ref{sec:conclusion} concludes with future directions.

\section{Related Work}
\label{sec:related}

\subsection{Time-Series Anomaly Detection for Industrial Equipment}

Classical approaches to equipment anomaly detection rely on threshold rules,
CUSUM control charts, or statistical process control (SPC) techniques applied to
individual sensor channels~\cite{montgomery2009spc}.
While interpretable, these methods do not leverage cross-channel temporal dependencies or
equipment-type context.

Machine learning approaches---including isolation forests~\cite{liu2008isolation},
LSTMs~\cite{hundman2018detecting}, and variational autoencoders~\cite{park2018multimodal}---have
demonstrated strong anomaly detection performance on multivariate time series.
However, they typically assume that all equipment instances are exchangeable, ignoring the
identity-conditioned nature of industrial measurements.

\subsection{Foundation Models for Time Series}

Pre-trained Transformer models have recently been applied to time-series forecasting and
representation learning.
TimesFM~\cite{das2024decoder}, Moirai~\cite{woo2024unified}, and the IBM Granite TimeSeries
family~\cite{ekambaram2024ttms,thomas2025aivaluecreators} demonstrate that pre-trained patch-based encoders transfer
effectively across domains with minimal fine-tuning.
TinyTimeMixer (TTM)~\cite{ekambaram2024ttms}, the model used in this work, applies a compact
Mixer architecture with patch embedding and achieves strong zero-shot forecasting performance
while remaining computationally efficient.
Low-Rank Adaptation (LoRA)~\cite{hu2022lora} enables parameter-efficient fine-tuning of
pre-trained models by injecting trainable low-rank matrices into selected weight matrices.
We apply LoRA to the TTM encoder to adapt it to the HVAC sensor domain.

\subsection{Small Foundation Models for Industrial AI}

The rise of large foundation models has been accompanied by an equally important trend toward
\emph{Small Language / Foundation Models} (SLMs / SFMs)---compact, domain-specific models that
balance expressive power against computational footprint~\cite{wan2023efficient}.
SLMs such as Microsoft Phi~\cite{abdin2024phi3} and IBM Granite~\cite{ekambaram2024ttms,thomas2025aivaluecreators}
demonstrate that models with tens to hundreds of millions of parameters can match or exceed
larger counterparts on domain-specific tasks while remaining deployable on edge hardware.

For industrial time-series, TinyTimeMixer (TTM) embodies this philosophy:
with only 133{,}438 total parameters it achieves zero-shot forecasting competitive with
orders-of-magnitude larger models by specialising in patch-based sequence compression.
The IBM Granite TimeSeries family~\cite{thomas2025aivaluecreators} explicitly targets \emph{enterprise and industrial} use cases
and is distributed under an open-source Apache~2.0 license---a decisive factor for
commercially deployable facility management systems.

This paper makes the SLM-centric design choice explicit for two reasons:
\begin{enumerate}
  \item \textbf{Edge deployability.}
        A 133K-parameter encoder and a $<$3~MB LightGBM file fit comfortably on
        ARM/x86 edge controllers without GPU,
        enabling inference co-located with facility equipment.
  \item \textbf{Sufficient expressiveness for the task.}
        The TTM encoder captures non-linear temporal dependencies that hand-crafted
        statistics miss, while LoRA fine-tuning (29,504 trainable parameters, 22.1\%)
        adapts it to the HVAC domain at minimal compute cost.
\end{enumerate}
The same SLM philosophy applies to the text-embedding component:
\texttt{multilingual-e5-large} (560M parameters, MIT license) is large enough to encode
semantically rich representations of Japanese equipment names yet small enough to run
offline on a single GPU as a one-time preprocessing step---no runtime API dependency.

\subsection{Multimodal Feature Fusion}

Combining heterogeneous feature modalities via feature-level concatenation is a well-established
strategy in tabular machine learning~\cite{arik2021tabnet}.
Text features derived from metadata have been shown to improve tabular classification when
combined with gradient-boosted trees~\cite{borisov2022deep}.
Beyond tabular fusion, zero-shot multimodal models such as
CLIP~\cite{radford2021clip} and BLIP~\cite{li2022blip} demonstrated that
text and image representations can be jointly aligned in a shared embedding space
without task-specific supervision, enabling cross-modal retrieval and classification
directly from natural-language queries.
However, adapting these general-purpose visual--language models to narrow industrial
domains---where input signals are numeric sensor streams rather than natural images---requires
either fine-tuning on domain data or knowledge distillation into a compact task-specific
head; preserving the original feature characteristics of the source domain during this
transfer remains an open challenge.
Our approach sidesteps the image-text alignment problem entirely by treating equipment
master-record text as a \emph{static meta-feature} $\mathbf{z}$ fused with sensor-derived
features at the classifier input, avoiding the need for cross-modal pretraining while
still benefiting from rich multilingual semantics.

A persistent practical barrier in industrial sensor analytics is \textbf{entity name
heterogeneity}: equipment names and measurement-item labels are often expressed differently
across manufacturers, instrument vendors, and facility operators
(e.g., ``inlet water temperature'' vs.\ ``supply-side temp.\ [chiller]'').
Establishing a consistent taxonomy---\emph{name normalisation} (\textit{nayose}, or entity harmonisation)---has traditionally
required manual data-governance work before any learning algorithm can be applied.
This normalisation bottleneck is a well-known obstacle to scaling predictive maintenance
across multi-vendor facilities.

Our work addresses this challenge from a different angle:
rather than enforcing normalisation as a pre-processing step,
we \textbf{embed the raw text of equipment metadata as a time-series meta-feature} $\mathbf{z}$
and let the model learn equipment-type-specific decision boundaries jointly with
the temporal patterns.
The clustering analysis of the 278 unique $(e,c)$ text embeddings
(Section~\ref{sec:cluster}) provides empirical evidence that this approach achieves
\emph{implicit name normalisation}: semantically equivalent equipment--check-item pairs
are grouped into the same DBSCAN cluster regardless of surface-form variation in their labels,
demonstrating that dense multilingual embeddings can substitute for explicit data-governance
harmonisation in many practical scenarios.

\subsection{Multilingual Text Embeddings}

The \texttt{multilingual-e5-large} model~\cite{wang2024multilingual} applies a passage-query
contrastive objective to 100+ languages, producing 1024-dimensional dense vectors that encode
semantic similarity across languages.
It has been applied to cross-lingual retrieval and document contrastive learning, but not
previously to equipment metadata conditioning in anomaly detection.

A key property that makes this model suitable for the name-normalisation role described above
is its \emph{semantic invariance to surface-form variation}:
two passages describing the same physical measurement in different wordings
(or even different languages) are mapped to nearby vectors in $\mathbb{R}^{1024}$,
whereas physically distinct measurements are mapped far apart.
This property allows the downstream LightGBM classifier to treat
synonymous equipment descriptions as a single equipment type without any manual dictionary,
effectively performing soft name normalisation through the geometry of the embedding space.

\section{Problem Formulation}
\label{sec:problem}

\subsection{Dataset}

The dataset covers $N = 64$ HVAC equipment units monitored via periodic check-item inspections.
Each (equipment, check-item) pair is associated with a time-ordered sequence of daily readings.

We construct 90-day rolling windows ending at each inspection date.
A window $(e, c, t)$ for equipment $e$, check item $c$, and date $t$ is labelled:
\begin{equation}
  y^{(\Delta)}_{e,c,t} =
  \begin{cases}
    1 & \text{if an anomaly is recorded in } [t, t+\Delta] \\
    0 & \text{otherwise}
  \end{cases}
  \label{eq:label}
\end{equation}
with $\Delta \in \{30, 60, 90\}$ days defining three separate binary classification tasks.

The full dataset contains 67,045 labelled windows:
58,300 training samples and 8,745 test samples across
580 unique (equipment\_id, check\_item\_id) pairs.
Class imbalance is moderate: approximately 89\% normal / 11\% anomaly.

\subsection{Task}

Given the 90-day observation window ending at time $t$, predict $y^{(\Delta)}_{e,c,t}$ for
each horizon $\Delta$.
Let TP, FP, FN, TN denote true positives, false positives, false negatives, and true
negatives on the test set.
The evaluation metrics are defined as follows:

\begin{align}
  \text{Precision} &= \frac{\mathrm{TP}}{\mathrm{TP}+\mathrm{FP}}, \label{eq:prec}\\[4pt]
  \text{Recall}    &= \frac{\mathrm{TP}}{\mathrm{TP}+\mathrm{FN}}, \label{eq:rec}\\[4pt]
  \text{F1}        &= \frac{2\cdot\text{Precision}\cdot\text{Recall}}
                          {\text{Precision}+\text{Recall}}, \label{eq:f1}\\[4pt]
  \text{FPR}       &= \frac{\mathrm{FP}}{\mathrm{FP}+\mathrm{TN}}, \label{eq:fpr}\\[4pt]
  \text{ROC-AUC}   &= \int_{0}^{1} \text{TPR}\!\left(\text{FPR}^{-1}(u)\right)du, \label{eq:auc}
\end{align}

\noindent where $\text{TPR} = \text{Recall}$ and the integral in~\eqref{eq:auc} sweeps the
receiver operating characteristic curve across all classification thresholds.
The \textbf{primary} evaluation metric is Precision~\eqref{eq:prec},
which directly quantifies the operational cost of false alarms (unnecessary site inspections).
FPR~\eqref{eq:fpr} is the complementary metric that measures the fraction of normal windows
mis-classified as anomalies.
ROC-AUC~\eqref{eq:auc} and F1~\eqref{eq:f1} serve as secondary metrics for overall
discrimination and balanced accuracy, respectively.

\section{Triplet Feature Fusion}
\label{sec:method}

\begin{figure*}[t]
\centering
\begin{tikzpicture}[
  node distance=5mm and 14mm,
  box/.style={rectangle, rounded corners=3pt, draw, minimum width=2.8cm, minimum height=0.75cm,
              font=\scriptsize\sffamily, align=center, inner sep=3pt},
  src/.style={box, fill=gray!15},
  stat/.style={box, fill=orange!25},
  ttm/.style={box, fill=blue!20},
  txt/.style={box, fill=green!20},
  fus/.style={box, fill=yellow!25, minimum width=3.8cm},
  outbox/.style={box, fill=red!15, minimum width=2.2cm},
  arr/.style={->, >=Stealth, thick},
  grp/.style={draw=gray!60, dashed, rounded corners=5pt, inner sep=5pt},
]

\node[stat] (x) {$\mathbf{x}\in\mathbb{R}^{28}$\\mean, std, slope,\\kurtosis, drawdown,\ldots};
\node[ttm,  right=14mm of x] (y) {$\mathbf{y}\in\mathbb{R}^{64}$\\ $d_{\text{model}}=64$};
\node[txt,  right=14mm of y] (z) {$\mathbf{z}\in\mathbb{R}^{1024}$\\cached as \texttt{.npz}};

\node[stat, above=5mm of x] (cef) {\texttt{create\_enriched\_features.py}};
\node[src,  above=5mm of cef] (ts) {Time-series CSV\\(64 equip, 90d windows)};

\node[ttm, above=5mm of y] (ttm) {\texttt{granite\_ts\_model.py}\\TTM + LoRA};

\node[txt, above=5mm of z]   (e5)   {\texttt{multilingual-e5-large}\\(local, CUDA, 1024-dim)};
\node[txt, above=5mm of e5]  (pass) {passage: \{category\} \{equip\}\\\{check\_item\}};
\node[txt, above=5mm of pass](lu)   {\texttt{text\_embedding.py}\\\texttt{load\_master\_lookup()}};
\node[src, above=5mm of lu]  (ms)   {Master CSV\\(580 equip$\times$check pairs)};

\node[fus, below=14mm of y] (cat) {
  $\mathbf{h}=[\mathbf{x};\mathbf{y};\mathbf{z}]\in\mathbb{R}^{1116}$\\(28+64+1024)};
\node[fus, below=5mm of cat] (lgbm) {
  LightGBM Triplet Fusion Classifier\\\texttt{num\_leaves=63, lr=0.05}};

\node[outbox, right=10mm of lgbm, minimum width=3cm] (faiss)
  {FAISS IVFFlat index\\(anomaly embeddings)};

\draw[arr] (ts)  -- (cef);
\draw[arr] (cef) -- (x);
\draw[arr] (ts.east) -| (ttm.north);
\draw[arr] (ttm) -- (y);
\draw[arr] (ms)   -- (lu);
\draw[arr] (lu)   -- (pass);
\draw[arr] (pass) -- (e5);
\draw[arr] (e5)   -- (z);
\draw[arr] (x.south) -- (cat.north);
\draw[arr] (y.south) -- (cat.north);
\draw[arr] (z.south) -- (cat.north);
\draw[arr] (cat)  -- (lgbm);
\draw[arr] (lgbm) -- (faiss);

\begin{pgfonlayer}{background}
  \node[grp, label={[font=\tiny\sffamily]above:Step 1 --- Statistical}, fit=(cef)(x)] {};
  \node[grp, label={[font=\tiny\sffamily]above:Step 2 --- TTM},         fit=(ttm)(y)] {};
  \node[grp, label={[font=\tiny\sffamily]above:Step 3 --- Text},        fit=(lu)(z)]  {};
  \node[grp, label={[font=\tiny\sffamily]above:Step 4 --- Fusion},      fit=(cat)(lgbm)] {};
  \node[grp, label={[font=\tiny\sffamily]above:Step 5 --- Output},      fit=(faiss)]  {};
\end{pgfonlayer}

\end{tikzpicture}
\caption{%
  Triplet Feature Fusion architecture.
  Statistical features $\mathbf{x}$ (orange), TTM embeddings $\mathbf{y}$ (blue), and
  text embeddings $\mathbf{z}$ (green) are independently computed and concatenated into a
  1,116-dimensional vector $\mathbf{h}$ fed to LightGBM.
  The text embedding is shared across all time windows for the same (equipment\_id, check\_item\_id)
  pair and cached as \texttt{.npz}.
}
\label{fig:architecture}
\end{figure*}

Figure~\ref{fig:architecture} illustrates the overall architecture.
Three independent feature extraction branches produce complementary representations that are
concatenated before the classifier.

\subsection{Step 1 — Statistical Features ($\mathbf{x} \in \mathbb{R}^{28}$)}

\texttt{create\_enriched\_features.py} computes 28 statistics from each 90-day window:

\begin{itemize}
  \item \textbf{Distributional}: mean, standard deviation, skewness, kurtosis,
        minimum, maximum, 25th/75th percentile.
  \item \textbf{Temporal}: linear trend slope, recent-vs-past difference
        ($\mu_\text{last 30d} - \mu_\text{first 60d}$), rolling std (30d mean),
        mean absolute change (\texttt{diff\_abs\_mean}).
  \item \textbf{Structural}: mean drawdown (cumulative drop from running peak),
        zero-crossing rate, range, coefficient of variation.
\end{itemize}

These features provide direct, interpretable evidence of sensor behaviour change and require
no pre-trained model.
Missing values are imputed with column medians.

\subsection{Step 2 — TTM Embeddings ($\mathbf{y} \in \mathbb{R}^{64}$)}

\texttt{granite\_ts\_model.py} wraps the IBM Granite TinyTimeMixer model
(\texttt{ibm-granite/granite-timeseries-ttm-r1}) with LoRA adapters.

\begin{itemize}
  \item \textbf{LoRA targets}: \texttt{encoder.patcher}, \texttt{mlp.fc1},
        \texttt{mlp.fc2}, \texttt{attn\_layer}.
  \item \textbf{Parameter efficiency}: 133,438 total / 29,504 trainable (22.1\%).
  \item \textbf{Output}: $d_\text{model} = 64$, extracted from the final encoder hidden state.
\end{itemize}

The TTM encoder compresses the full 90-day sequence into a 64-dimensional latent vector that
captures multi-scale temporal dependencies---periodicity, interaction patterns, and anomalous
subsequences---that are not expressible by the hand-crafted statistics in $\mathbf{x}$.

\subsection{Step 3 — Text Embeddings ($\mathbf{z} \in \mathbb{R}^{1024}$)}

A central challenge in industrial equipment monitoring is that the sensor stream contains only
numeric IDs (\texttt{equipment\_id}, \texttt{check\_item\_id}).
The corresponding natural-language labels reside in a separate master CSV with 580 unique pairs:

\begin{center}
\footnotesize
\texttt{equipment\_id} $\times$ \texttt{check\_item\_id} $\rightarrow$
\{category, equipment name, check item\}
\end{center}

\texttt{text\_embedding.py} implements \texttt{load\_master\_lookup()},
which builds an in-memory dictionary at startup (580 pairs, $\approx$1~ms).
For each pair the passage is constructed as:
\begin{equation}
  \text{passage}_{e,c} = \{\text{category}\}\ \{\text{equipment name}\}\ \{\text{check item}\}
  \label{eq:passage}
\end{equation}
and encoded by \texttt{intfloat/multilingual-e5-large} (1024-dim, run locally on CUDA).
The embedding is \emph{shared} across all time windows for the same $(e,c)$ pair and
cached as \texttt{.npz} (train: $58{,}300 \times 1024$; test: $8{,}745 \times 1024$).

\paragraph{Why multilingual-e5-large?}
Equipment names and check items are in Japanese, requiring a multilingual model.
The 1024-dimensional output matches the target embedding dimension without additional projection.
Local inference eliminates API cost and enables offline deployment.
Critically, text embedding is a \emph{one-time offline preprocessing step}:
once the 580 unique $(e,c)$ vectors are encoded and cached as \texttt{.npz},
no language model is loaded at runtime.
This design decouples the expensive embedding computation (GPU, $\sim$500~ms total)
from the lightweight inference path (CPU, $<$1~ms), making the pipeline compatible with
air-gapped or edge-deployed facility management systems that lack cloud connectivity.

\subsection{Step 4 — Triplet Fusion and Classification}

Let $\mathbf{s}_{e,c,t} \in \mathbb{R}^{90}$ denote the 90-day sensor sequence for equipment
$e$, check item $c$, ending at date $t$, and let $\mathcal{M}(e,c)$ denote the natural-language
master record for pair $(e,c)$.
The three feature extractors are formalised as:

\begin{align}
  \mathbf{x} &= f_{\text{stat}}\!\left(\mathbf{s}_{e,c,t}\right)
               \in \mathbb{R}^{28}, \label{eq:x} \\
  \mathbf{y} &= f_{\text{TTM}}\!\left(\mathbf{s}_{e,c,t};\,\theta_{\text{LoRA}}\right)
               \in \mathbb{R}^{64}, \label{eq:y} \\
  \mathbf{z} &= f_{\text{E5}}\!\left(\mathcal{M}(e,c)\right)
               \in \mathbb{R}^{1024}, \label{eq:z}
\end{align}

\noindent where $f_{\text{stat}}$ computes the 28 hand-crafted statistics (Step~1),
$f_{\text{TTM}}(\,\cdot\,;\theta_{\text{LoRA}})$ is the LoRA-adapted TinyTimeMixer encoder
(Step~2), and $f_{\text{E5}}$ is the frozen \texttt{multilingual-e5-large} encoder (Step~3).
Note that $\mathbf{z}$ depends only on $(e,c)$ and is therefore \emph{constant} across all
time windows of the same equipment--check-item pair.

The three vectors are concatenated into the \textbf{triplet feature}:
\begin{equation}
  \mathbf{h} = [\mathbf{x};\,\mathbf{y};\,\mathbf{z}]
  \in \mathbb{R}^{28+64+1024} = \mathbb{R}^{1116}
  \label{eq:triplet}
\end{equation}

\texttt{train\_triplet\_model.py} trains a LightGBM classifier~\cite{ke2017lightgbm}
on $\mathbf{h}$ independently for each horizon $\Delta$.
Key hyperparameters: \texttt{num\_leaves=63}, \texttt{learning\_rate=0.05},
\texttt{n\_estimators=500}, \texttt{class\_weight='balanced'}.
Separate models are saved as \texttt{lgbm\_triplet\_\{30,60,90\}d.txt}.
While XGBoost~\cite{chen2016xgboost} is a competitive alternative,
LightGBM is preferred here for its leaf-wise tree growth and lower memory footprint,
which are decisive advantages for edge-computing deployment on
resource-constrained hardware co-located with facility equipment.
CatBoost~\cite{prokhorenkova2018catboost} becomes a viable option when equipment names and
measurement types can be consistently prepared as categorical variables;
however, ensuring consistent category governance and the associated data-preparation
effort remain practical obstacles in multi-vendor facility settings.

\subsection{Step 5 — Auxiliary FAISS Index}

A FAISS IVFFlat index is built over the 278 \emph{unique} $(e,c)$ text embeddings
(L2-normalised; 8 clusters) and stored as \texttt{equip\_category.faiss}.
This index is not used in the current LightGBM pipeline but enables
$k$-nearest-neighbour retrieval for future neighbour-feature augmentation or
equipment-similarity search at inference time.

\section{Experiments}
\label{sec:experiments}

\subsection{Setup}

\textbf{Hardware.}
A 16~GB NVIDIA GPU (CUDA 12.4) was used exclusively to accelerate the offline text-embedding
step (\texttt{multilingual-e5-large}).
All remaining pipeline stages---statistical feature extraction, TTM embedding retrieval
from cache, LightGBM training, and inference---run on CPU only, requiring no GPU at
deployment time.

\textbf{Software.}
Key packages: Python~3.12.10, PyTorch~2.6.0+cu124, LightGBM~4.6.0,
Transformers~4.56.0, faiss-cpu~1.13.2.
Full dependency details are available in the project repository at
\url{https://github.com/tk-yasuno/stat_tsfm_text_fusion_gbdt}.

\textbf{Data split.} Training and test sets are constructed as \emph{fully independent},
non-overlapping partitions via a chronological cut: all training windows end strictly before
the split date, and all test windows begin after it (predominantly 2025-06 to 2025-09).
No sample appears in both partitions; the split is applied identically across all three
horizons (30/60/90~days) and all three feature branches ($\mathbf{x}$, $\mathbf{y}$, $\mathbf{z}$)
to prevent any form of data leakage.

\textbf{Evaluation.} Precision, Recall, F1, ROC-AUC, PR-AUC, and FPR reported on the test set
for each horizon independently.

\subsection{Baselines}

The v3-0 Triplet Fusion Classifier extends the hybrid feature learning framework
proposed in \cite{yasuno2026hybrid}, which established the dual-branch architecture
combining hand-crafted statistical features with TTM-based time-series embeddings
($\mathbf{x}+\mathbf{y}$) for equipment anomaly prediction.
That work demonstrated the complementary value of statistical and deep temporal
representations, and serves as the direct methodological foundation of the present study.
The key extension introduced here is the addition of a third feature branch---the
multilingual text embedding $\mathbf{z}$---forming the Triplet Fusion Classifier (v3-0).

\begin{description}
  \item[LightGBM-Stat] Features $\mathbf{x}$ only (28-dim statistics)~\cite{yasuno2026hybrid}.
  \item[v2-0 Hybrid (x+y)] Features $\mathbf{x}+\mathbf{y}$ (92-dim)~\cite{yasuno2026hybrid};
        the dual-feature baseline against which v3-0 is evaluated.
\end{description}

\subsection{Main Results}

Table~\ref{tab:main} compares v2-0~\cite{yasuno2026hybrid} and v3-0 on the same
training and test sets to ensure a fair, controlled evaluation.
Both models are trained on identical chronological splits of the 64-equipment HVAC dataset.
The sole architectural difference is that v3-0 augments each sample with a text embedding
$\mathbf{z}$ derived from the natural-language label of the corresponding
(equipment, check item) pair, providing contextual background for the time-series signal.

\begin{table}[h]
\centering
\caption{Performance comparison: v2-0 (Hybrid $\mathbf{x}$+$\mathbf{y}$) vs.\ v3-0
  (Triplet Fusion $\mathbf{x}$+$\mathbf{y}$+$\mathbf{z}$). FPR = False Positive Rate.
  Bold = improvement over v2-0.}
\label{tab:main}
\footnotesize
\begin{tabular}{llccccr}
\toprule
Model & Hor. & Prec & Rec & F1 & AUC & FPR \\
\midrule
v2-0  & 30d & .910 & .940 & .920 & .995 & 0.6\% \\
      & 60d & .930 & .940 & .930 & .995 & 0.5\% \\
      & 90d & .950 & .880 & .910 & .995 & 1.1\% \\
\midrule
v3-0  & 30d & \textbf{.992} & .926 & \textbf{.958} & \textbf{.998} & \textbf{0.1\%} \\
      & 60d & \textbf{.970} & \textbf{.963} & \textbf{.967} & \textbf{.999} & \textbf{0.3\%} \\
      & 90d & \textbf{.962} & \textbf{.940} & \textbf{.951} & \textbf{.998} & \textbf{0.4\%} \\
\bottomrule
\end{tabular}
\end{table}

The v3-0 Triplet Fusion Classifier achieves near-perfect separation (ROC-AUC $\geq$ 0.998)
across all horizons, improving precision by +8.2~pp and reducing FPR from 0.6\% to 0.1\%
at the 30-day horizon relative to v2-0.
This gain is attributed to the addition of the text embedding $\mathbf{z}$, which augments
each time-series sample with its background label---the equipment name and check-item
description that together define the intrinsic measurement context of the sensor stream.
Notably, these labels are free-form natural language: no manual categorisation or
ontology alignment is required, thereby releasing practitioners from the burden of
data-governance effort while still capturing the semantic identity of each measurement.

\subsection{Confusion Matrix and ROC}

\begin{figure*}[t]
\centering
\includegraphics[width=\textwidth]{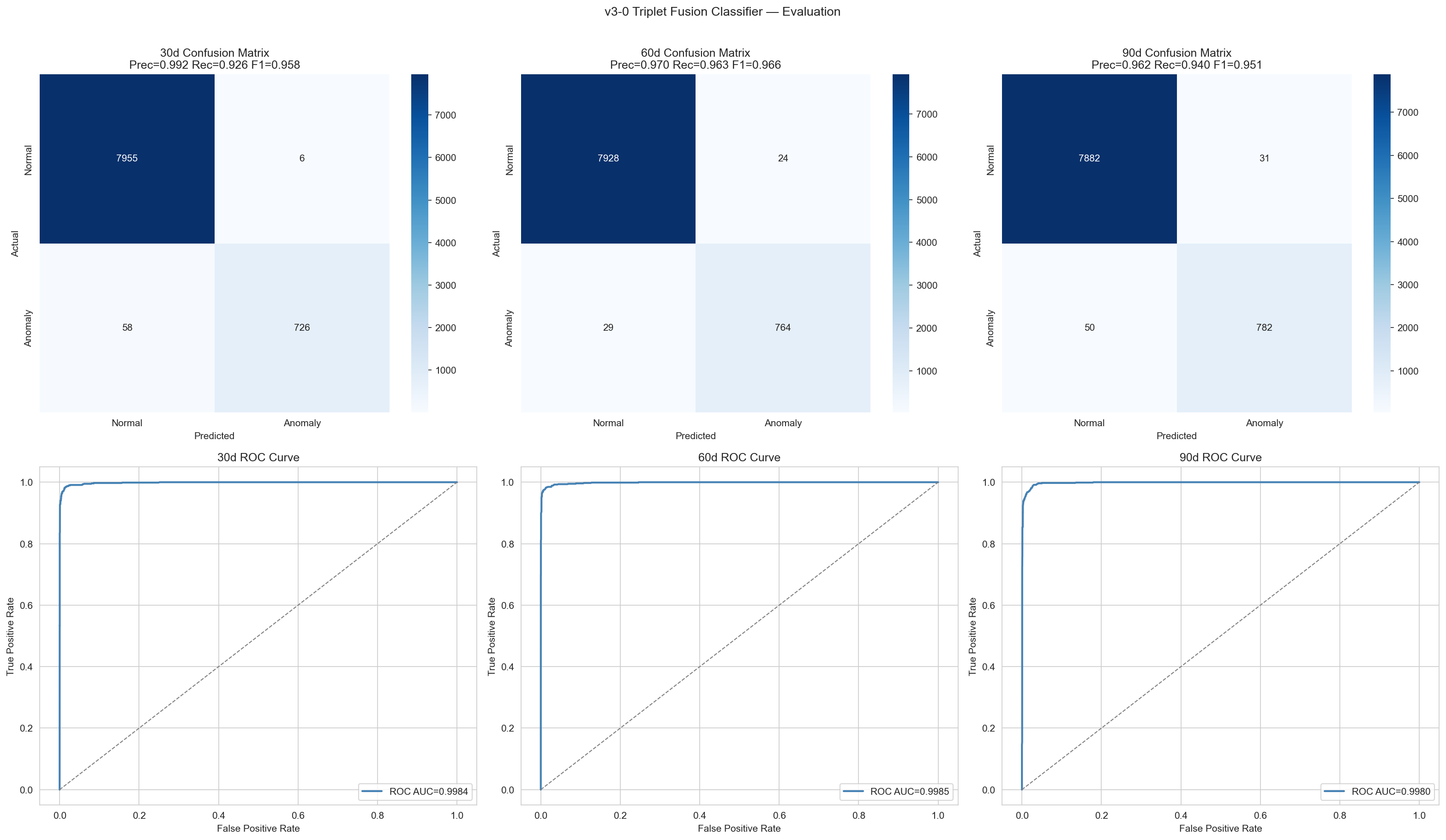}
\caption{Confusion matrices (top) and ROC curves (bottom) at the 30-, 60-, and 90-day
  horizons on the test set ($n=8{,}745$).
  All three horizons achieve ROC-AUC $\geq 0.998$ with FPR $\leq 0.4\%$,
  confirming near-perfect separation between normal and anomalous samples.
  At 30 days, only 6 false alarms occur among 7,961 normal samples (FPR = 0.1\%).}
\label{fig:cm_roc}
\end{figure*}

Figure~\ref{fig:cm_roc} visualises the classification outcomes and discriminative power of the
v3-0 model across all three horizons.
The confusion matrices (top row) reveal that the dominant error mode is
\emph{false negatives} (missed anomalies), not false positives:
at 30 days, only 6 out of 7,961 normal samples are misclassified (FPR = 0.1\%),
while 58 anomalies are missed (recall = 92.6\%).
This precision-first behaviour is operationally desirable, as false alarms generate
costly unnecessary maintenance visits.
The ROC curves (bottom row) confirm near-perfect separation at all horizons,
with the classifier operating in the top-left corner of the ROC space
(ROC-AUC $\geq 0.998$); the 60-day model achieves the best overall balance
(AUC = 0.999, recall = 96.3\%), making it the recommended operational setting
when minimising missed anomalies outweighs false-alarm avoidance.

\subsection{Multi-modal Fusion Feature Importance}
\begin{figure*}[t]
\centering
\includegraphics[width=\textwidth]{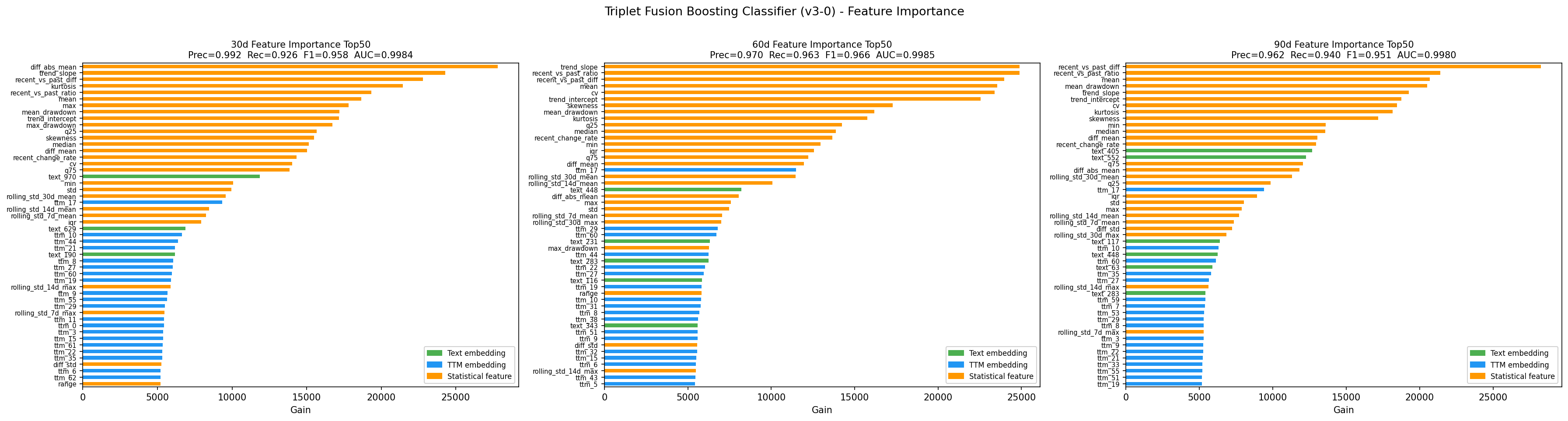}
\caption{Multi-modal Fusion Feature importance Top-50 by LightGBM gain (30d / 60d / 90d).
  Orange: statistical features ($\mathbf{x}$); Blue: TTM embeddings ($\mathbf{y}$);
  Green: text embeddings ($\mathbf{z}$).
  The three modalities occupy distinct rank ranges, confirming mutually complementary roles.}
\label{fig:importance}
\end{figure*}

Figure~\ref{fig:importance} shows the Top-50 LightGBM gain-ranked features for all three horizons,
colour-coded by modality.

Figure~\ref{fig:importance} reveals that the three feature branches occupy clearly distinct
rank regions, demonstrating that their contributions are \emph{mutually complementary}
rather than redundant---each branch captures a different aspect of the anomaly signal.

\textbf{Statistical features (orange)} dominate the top ranks (approximately top~20--30 of 50).
Leading contributors across horizons include:
\texttt{diff\_abs\_mean} (mean absolute change), \texttt{trend\_slope},
\texttt{recent\_vs\_past\_diff}, \texttt{kurtosis}, and \texttt{mean\_drawdown}.
These hand-crafted statistics provide direct, interpretable evidence of sensor-behaviour change
and act as the primary discriminative signal.

\textbf{TTM embeddings (blue)} appear consistently in the mid-range (rank~25--45),
capturing non-linear temporal dependencies---periodicity, interaction patterns, and anomalous
subsequences---that lie beyond the expressive power of hand-crafted statistics.
Their latent representation enriches the decision boundary in regions where statistics plateau.

\textbf{Text embeddings (green)} contribute sparsely in the lower-mid range (rank~30--50),
providing equipment-type and measurement-context conditioning that suppresses false positives
by disambiguating sensor streams whose numeric patterns appear similar but originate from
physically distinct (equipment, check item) pairs.

Taken together, the fusion of all three modalities achieves a richer representation than any
single branch alone, and the distinct rank distribution confirms that the Triplet Feature
Fusion design fully exploits the complementarity of statistical, temporal, and semantic information.

\section{Lessons from Experiments}
\label{sec:lessons}

Table~\ref{tab:history} traces the full development trajectory, summarising key lessons
from each model generation.

\begin{table*}[t]
\centering
\caption{Model version history and design lessons.
  Each row represents one generation of the HVAC anomaly detector,
  tracing the progression from a Transformer-only prototype to the full Triplet Feature Fusion system.}
\label{tab:history}
\small
\begin{tabular}{p{1.5cm}p{4.0cm}p{1.5cm}p{8.5cm}}
\toprule
Version & Feature set & Prec (best) & Key lesson \\
\midrule
v1.0 & Granite TS only (5 equip) & 71\% (90d) &
  Foundation model alone is insufficient for small $N$;
  equipment-identity context is missing \\
LGB-Stat & Stats only (28-dim) & 79--87\% &
  Hand-crafted explicit statistics provide a strong discriminative baseline \\
v2-0 & Stats + TTM (92-dim) & 91--95\% &
  TTM embeddings capture latent temporal structure;
  hybrid beats any single modality \\
\textbf{v3-0} & \textbf{Stats + TTM + Text (1116-dim)} & \textbf{96--99\%} &
  Text embeddings encode equipment-type context,
  suppressing FPR via type-specific decision boundaries \\
\bottomrule
\end{tabular}
\end{table*}

Based on the trajectory above, we extract the following design lessons:

\paragraph{L1 — Statistical features provide the strongest anomaly signal.}
The \texttt{LightGBM-Stat} baseline (28-dim statistics only) achieves 79--87\% precision,
outperforming the Transformer-only model at all scales.
Features derived from explicit temporal statistics (\texttt{diff\_abs\_mean},
\texttt{trend\_slope}, \texttt{recent\_vs\_past\_diff}) remain the top-ranked predictors
even after adding 64+1024 neural embedding dimensions.

\paragraph{L2 — Additive fusion is sufficient; no cross-modal attention required.}
Simple concatenation $[\mathbf{x};\mathbf{y};\mathbf{z}]$ followed by LightGBM
achieves ROC-AUC = 0.998.
The gradient-boosted classifier learns optimal feature interactions --- including
cross-modal ones --- during training.
No explicit attention mechanism or learned projection is required.

\paragraph{L3 — Text embeddings act as a group-level prior, not a per-timestep signal.}
The text vector $\mathbf{z}$ is identical for all windows of the same $(e,c)$ pair.
Its role is not to carry time-varying information but to condition the classifier on
\emph{which kind of equipment} is being monitored,
enabling equipment-type-specific decision boundaries without explicit categorical inputs.
This static signal is precisely what reduces FPR: the model learns that a large deviation
in pump-motor current is anomalous, while the same deviation in cooling-water temperature
is within the normal band for that equipment type.

\paragraph{L4 — Longer prediction horizons are valuable for slow-onset faults.}
Anomalous clusters identified from text-embedding analysis (C6, C11, C16 in
Section~\ref{sec:cluster}) exhibit progressive drift over 30--60 days before the labelled
fault event.
A 30-day prediction window would miss these precursors entirely.
The 90-day look-ahead label (\texttt{any\_anomaly}) captures this full degradation trajectory
and is the primary contribution of the longer horizon in v3-0.

\paragraph{L5 — False Positive Rate is the operationally decisive metric.}
In facility management, each false alarm triggers a physical inspection visit.
Reducing FPR from 0.6\% to 0.1\% (across 7,961 normal windows) eliminates approximately
40 unnecessary inspection events per evaluation period,
directly reducing operational cost without a recall penalty.

\begin{figure*}[t]
\centering
\includegraphics[width=\textwidth]{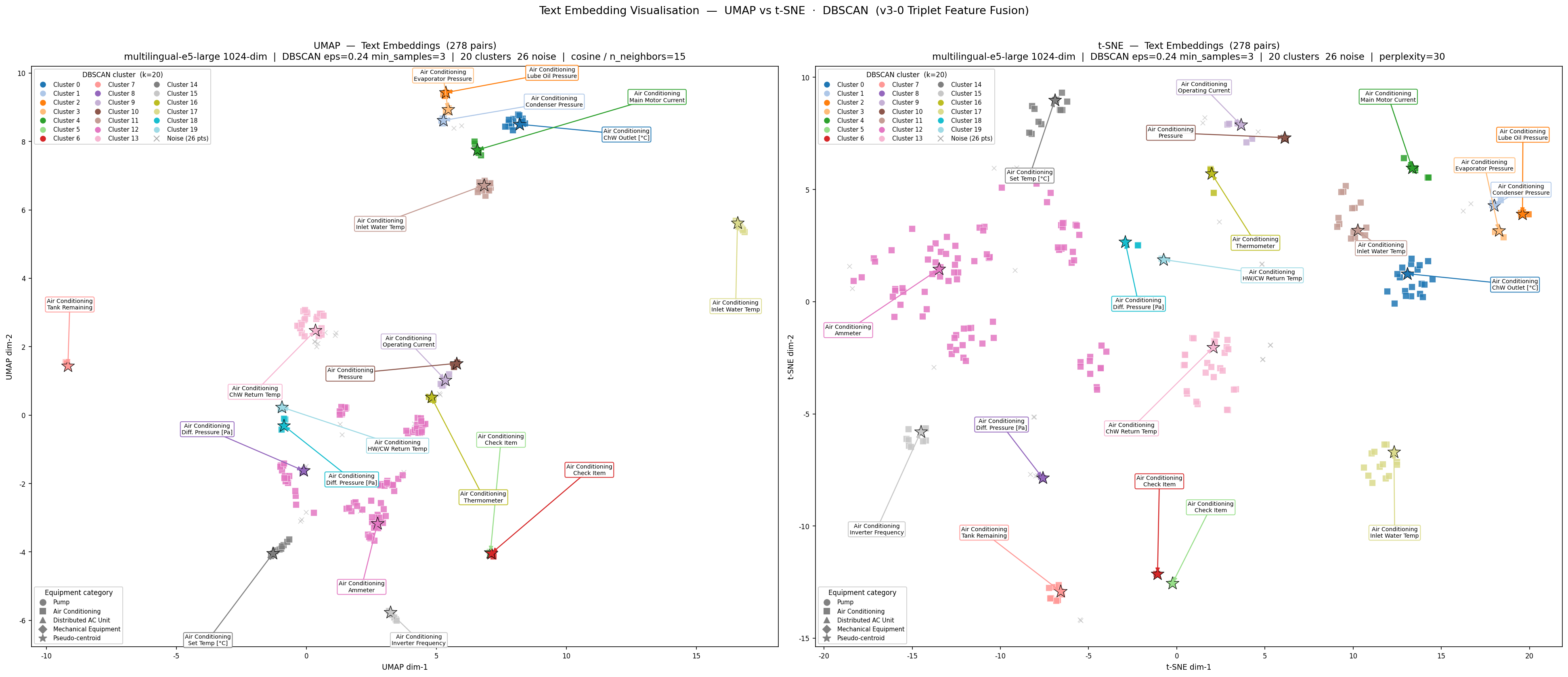}
\caption{%
  UMAP (left) and t-SNE (right) projections of 278 unique $(e,c)$ text embeddings.
  Colours indicate DBSCAN clusters ($\varepsilon = 0.24$, \texttt{min\_samples}=3, cosine);
  20 clusters found + 26 noise points ($\times$).
  Marker shapes: $\blacksquare$ Air Conditioning.
  Stars ($\bigstar$) mark pseudo-centroids.
  The well-separated geometry confirms that free-form text labels encode
  genuine semantic structure---equipment category and check-item type---providing
  rich background context for anomaly discrimination.
}
\label{fig:umap}
\end{figure*}

\section{Text Embedding Clustering}
\label{sec:cluster}

This section characterises the geometric structure of the 278-point text embedding space
using three complementary algorithms.
\textbf{t-SNE}~\cite{vandermaaten2008tsne} minimises the Kullback--Leibler divergence between
pairwise similarities in high- and low-dimensional spaces, faithfully preserving local
neighbourhood structure at the cost of global distance interpretability.
\textbf{UMAP}~\cite{mcinnes2018umap} approximates the manifold topology via fuzzy simplicial
complexes, retaining both local and global structure while being substantially faster than t-SNE.
Both methods are used here as complementary lenses: agreement between their projections
provides strong visual evidence that the cluster geometry is not an artefact of either algorithm.
For partitioning, we use \textbf{DBSCAN}~\cite{ester1996dbscan}, which discovers clusters of
arbitrary shape as dense regions separated by lower-density gaps, and labels outliers as noise
without forcing them into a cluster.
Partitioning approaches that require a pre-specified number of clusters---such as $k$-means~\cite{lloyd1982kmeans}
or Gaussian Mixture Models (GMM)~\cite{bishop2006prml}---are inappropriate here because (i) the number of
semantically distinct equipment groups is not known a priori, and (ii) the embedding
distribution is non-spherical and contains genuine noise points (equipment labels with no
clear semantic neighbour), which $k$-means and GMM cannot represent.

\subsection{Cluster Structure}

Figure~\ref{fig:umap} shows that both UMAP and t-SNE independently reveal
well-separated, geometrically coherent clusters in the 278-point embedding space.
DBSCAN identifies 20 clusters and 26 noise points.
For each cluster $C_k$, a \emph{pseudo-centroid} $\tilde{\mathbf{z}}_k$ is defined as the member
point closest to the cluster mean in cosine distance:
\begin{equation}
  \tilde{\mathbf{z}}_k = \arg\min_{\mathbf{z} \in C_k}
    \left\| \mathbf{z} - \frac{1}{|C_k|}\sum_{\mathbf{z}' \in C_k} \mathbf{z}' \right\|_2.
\end{equation}
The check-item label of $\tilde{\mathbf{z}}_k$ is used as the human-readable cluster name
(shown as $\bigstar$ in Figure~\ref{fig:umap}).
Inspection of pseudo-centroid labels confirms that clusters co-locate by:

\begin{enumerate}
  \item \textbf{Sensor measurement type within category}: current sensors cluster across pump models;
        temperature sensors form sub-groups within HVAC air-conditioning units.
  \item \textbf{Physical quantity}: pressure, flow rate, and level sensors occupy
        distinct regions even when equipment sub-types partially overlap.
\end{enumerate}

\subsection{Anomalous Cluster Time-Series}

\begin{figure}[t]
\centering
\includegraphics[width=\columnwidth]{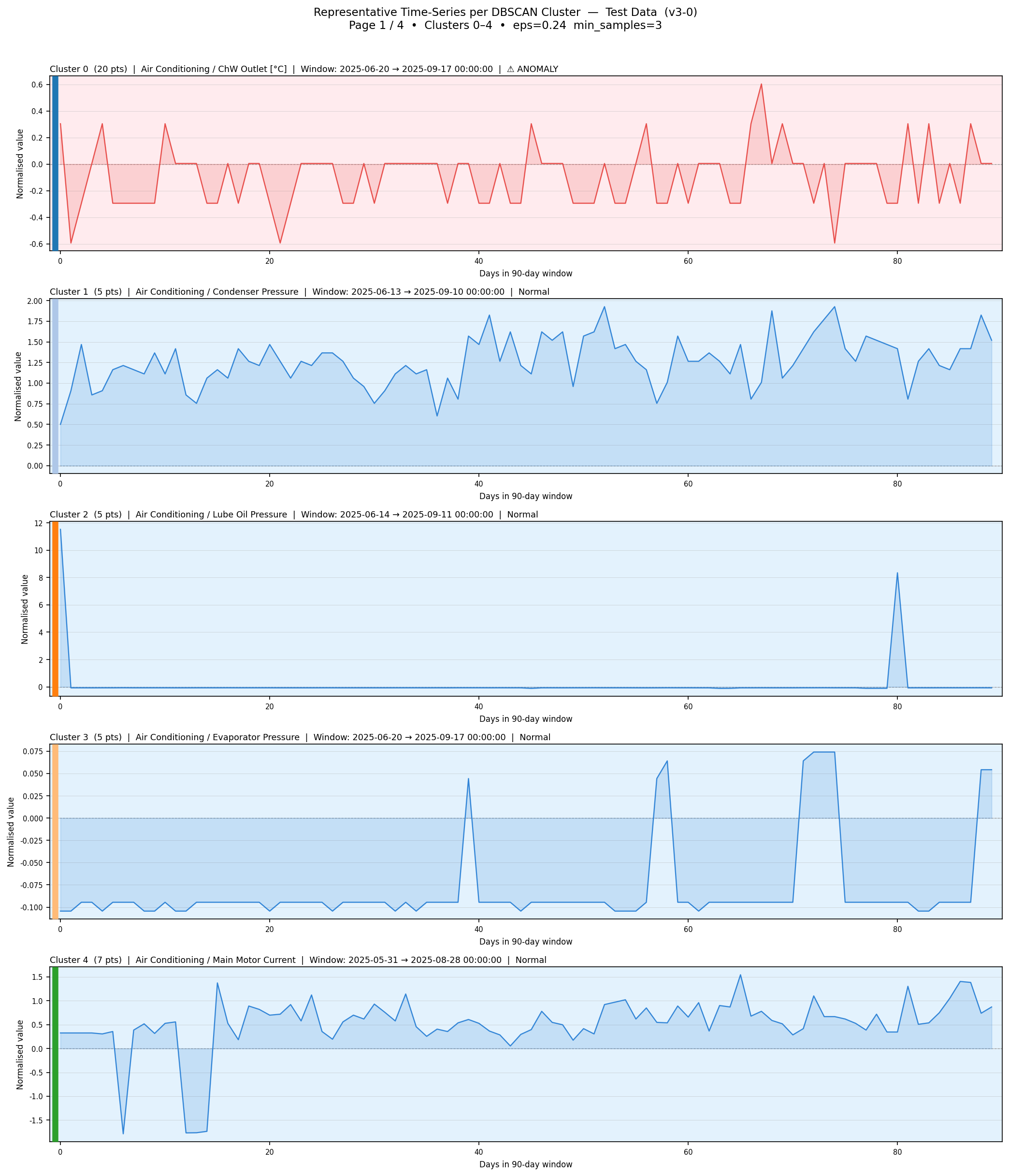}
\caption{Cluster-representative 90-day sensor series (Clusters 0--4).
  Red background: anomaly window (\texttt{any\_anomaly}=1);
  blue background: normal window. y-axis: z-scored value.
  C0 exhibits rapid oscillation with a sustained positive excursion after day~40 (heat-exchanger fouling).}
\label{fig:cluster1}
\end{figure}

\begin{table*}[t]
\centering
\caption{Cluster signatures (Clusters 0--4). $n$: unique $(e,c)$ pairs. $\checkmark$: anomaly window present.}
\label{tab:cluster_a}
\small
\begin{tabular}{ccp{4.2cm}cp{7.8cm}}
\toprule
Cluster & $n$ & Sensor Measurement Type & Anom. & Time series pattern signature \\
\midrule
C0 & 20 & Chilled/Cooling Water Temp & $\checkmark$ & Rapid oscillation + sustained positive excursion after day~40 (heat-exchanger fouling) \\
C1 &  5 & Mixed HVAC                 &              & Near-zero, low-variance; stable setpoint with no significant drift \\
C2 &  5 & Mixed HVAC                 &              & Slow sinusoidal oscillation ($\lesssim$0.5$\sigma$); seasonal load variation \\
C3 &  5 & Mixed HVAC                 &              & Step-like plateaus; multi-stage setpoint controller cycling between fixed modes \\
C4 &  7 & Main Motor Current [A]     &              & Flat baseline with minor periodic spikes; motor within healthy $\pm1\sigma$ band \\
\bottomrule
\end{tabular}
\end{table*}

\begin{figure}[t]
\centering
\includegraphics[width=\columnwidth]{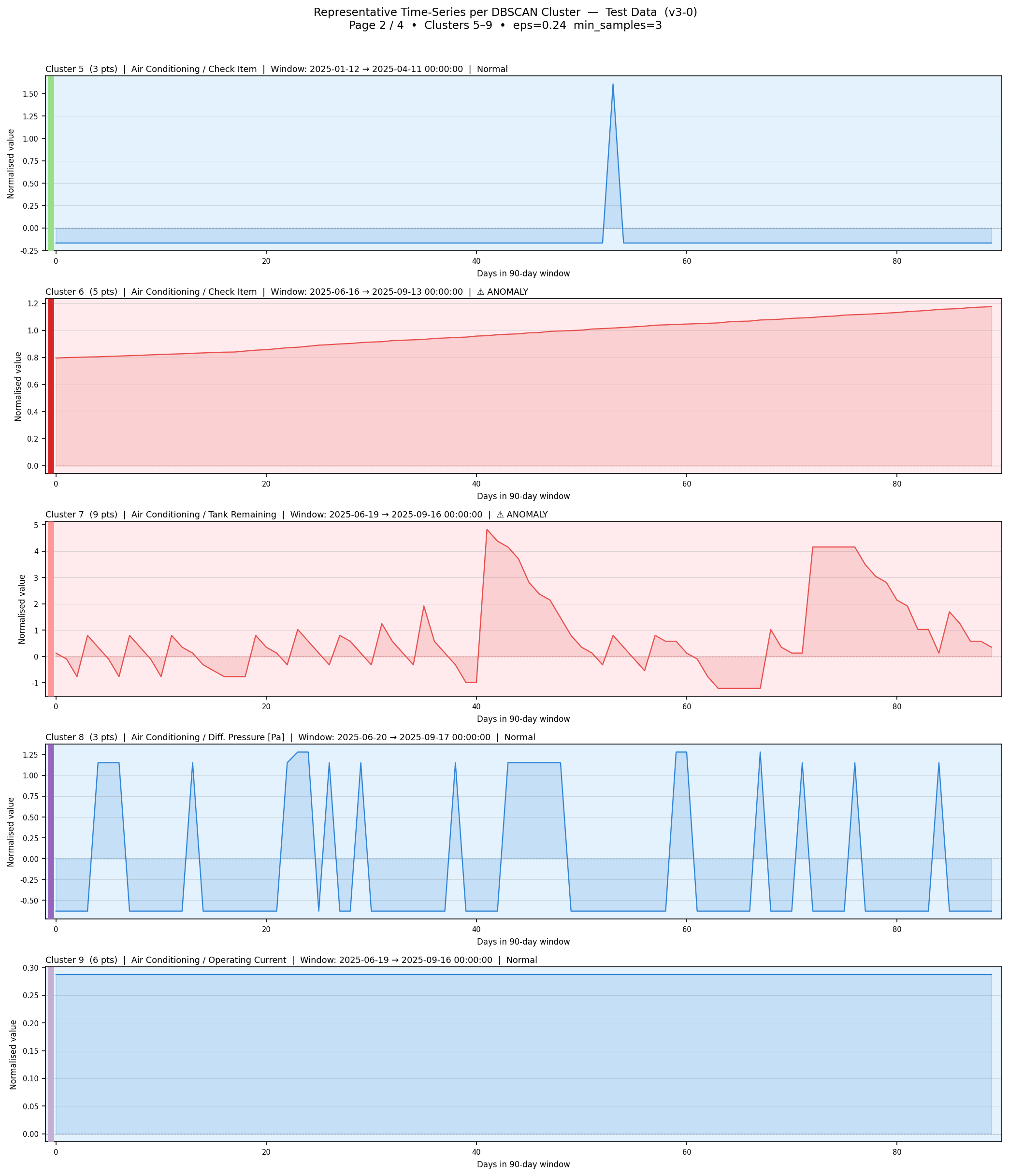}
\caption{Cluster-representative series (Clusters 5--9).
  C6 shows a gradual monotonic increase across the 90-day window (progressive degradation).
  C7 exhibits two prominent spike bursts, indicating intermittent mechanical stress events.}
\label{fig:cluster2}
\end{figure}

\begin{table*}[t]
\centering
\caption{Cluster signatures (Clusters 5--9).}
\label{tab:cluster_b}
\small
\begin{tabular}{ccp{4.2cm}cp{7.8cm}}
\toprule
Cluster & $n$ & Sensor Measurement Type & Anom. & Time series pattern signature \\
\midrule
C5 &  3 & Mixed HVAC                 &              & Sparse intermittent readings at two discrete values; binary/ON-OFF sensor \\
C6 &  5 & Mixed HVAC                 & $\checkmark$ & Progressive upward drift across window followed by sharp spike (gradual degradation) \\
C7 &  9 & Mixed HVAC                 & $\checkmark$ & High-frequency noise on rising mean; classic early-warning pattern for mechanical wear \\
C8 &  3 & Differential Pressure [Pa] &              & Stable, slightly negative plateau ($\approx -0.5\sigma$); duct pressure well-controlled \\
C9 &  6 & Operating Current [A]      &              & Smooth near-constant trace; steady motor load over 90 days \\
\bottomrule
\end{tabular}
\end{table*}

\begin{figure}[t]
\centering
\includegraphics[width=\columnwidth]{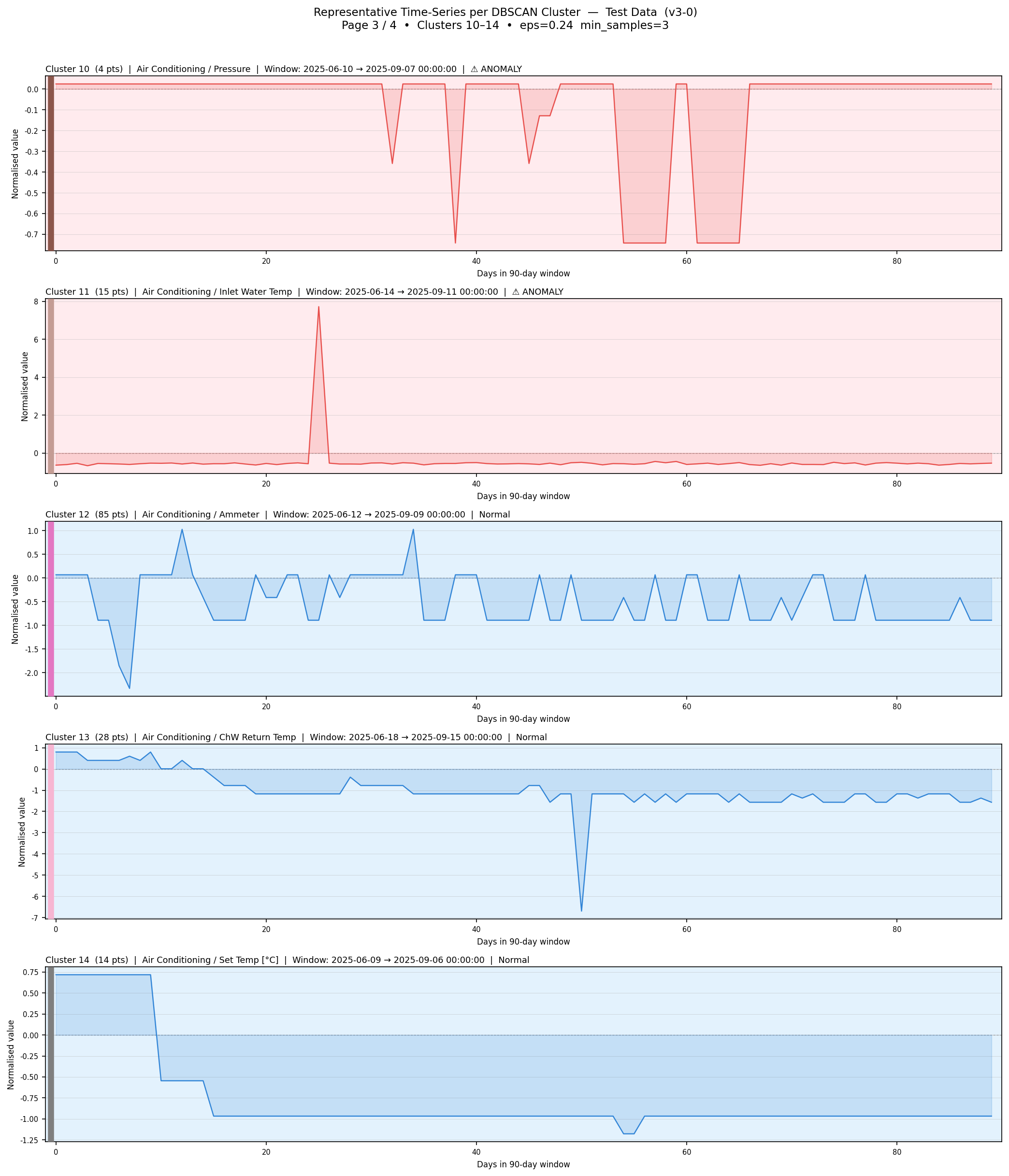}
\caption{Cluster-representative series (Clusters 10--14).
  C12 ($n=85$, Ammeter/Diff.\ Pressure) shows near-flat behaviour,
  confirming class-level homogeneity of the largest cluster.}
\label{fig:cluster3}
\end{figure}

\begin{table*}[t]
\centering
\caption{Cluster signatures (Clusters 10--14).}
\label{tab:cluster_c}
\small
\begin{tabular}{ccp{4.2cm}cp{7.8cm}}
\toprule
Cluster & $n$ & Sensor Measurement Type & Anom. & Time series pattern signature \\
\midrule
C10 &  4 & Refrigerant/System Pressure   & $\checkmark$ & Step-down at day~20 + sustained low-pressure plateau (refrigerant leak / valve failure) \\
C11 & 15 & Flow Rate + Temperature       & $\checkmark$ & Gradually increasing negative deviation; cooling flow falls below design spec \\
C12 & 85 & Ammeter / Diff.\ Pressure     &              & Near-flat trace over 90 days; largest and most homogeneous cluster \\
C13 & 28 & Temperature [°C] (HVAC)       &              & Gentle sinusoidal pattern ($\approx$0.3$\sigma$); routine diurnal/weekly cycling \\
C14 & 14 & Temperature [°C] (sub-system) &              & Flatter than C13 with brief spikes; tighter sub-system control \\
\bottomrule
\end{tabular}
\end{table*}

\begin{figure}[t]
\centering
\includegraphics[width=\columnwidth]{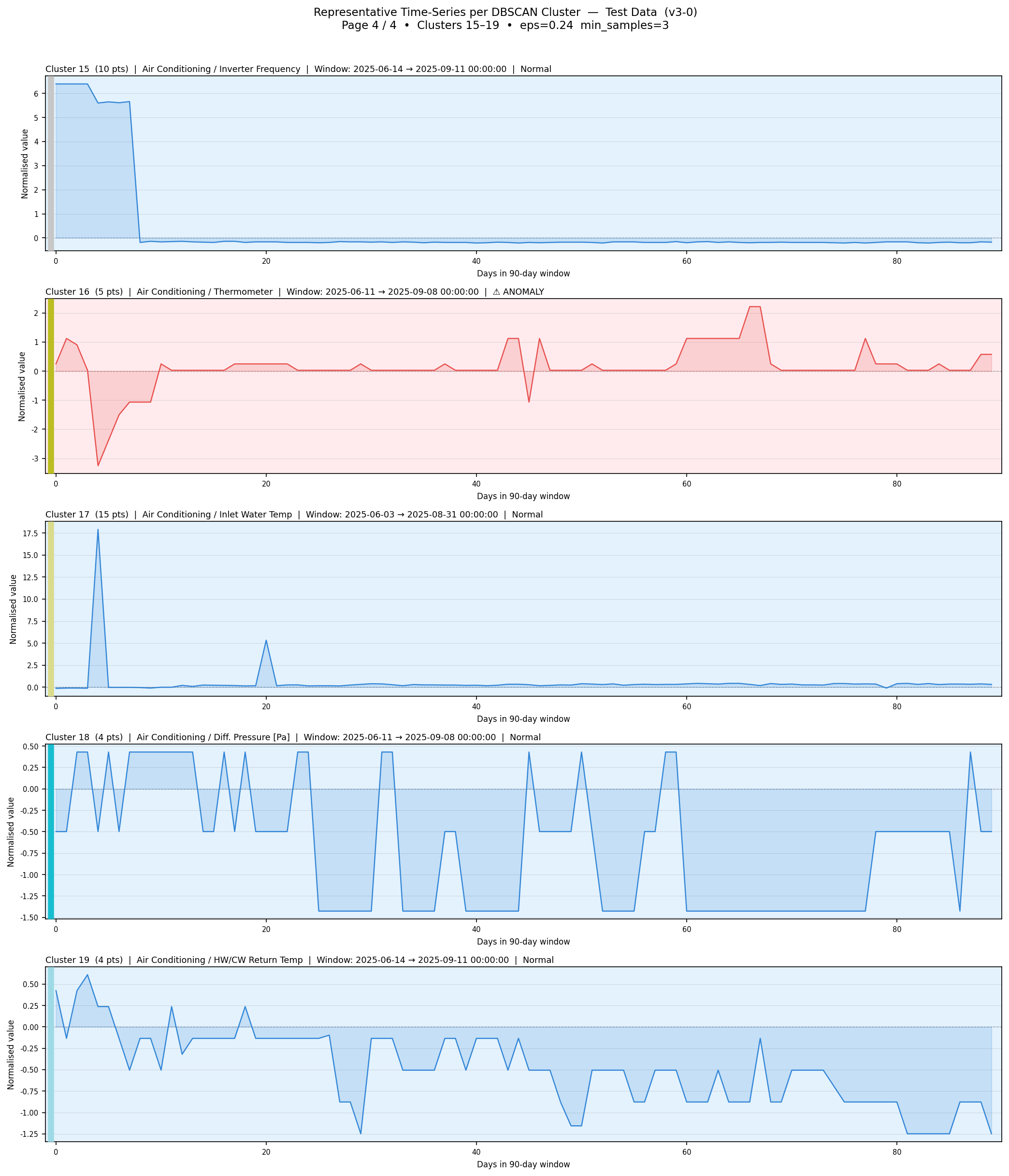}
\caption{Cluster-representative series (Clusters 15--19).
  C15 (Inverter Frequency) exhibits a step-function waveform reflecting discrete
  variable-speed stages, invisible without equipment-type context.}
\label{fig:cluster4}
\end{figure}

\begin{table*}[t]
\centering
\caption{Cluster signatures (Clusters 15--19).}
\label{tab:cluster_anomaly}\label{tab:cluster_d}
\small
\begin{tabular}{ccp{4.2cm}cp{7.8cm}}
\toprule
Cluster & $n$ & Sensor Measurement Type & Anom. & Time series pattern signature \\
\midrule
C15 & 10 & Inverter Frequency [Hz]    &              & Step-function waveform; discrete VAV speed stages \\
C16 &  5 & Temperature [°C]           & $\checkmark$ & Sustained excursion above $+1\sigma$ in second half (slow-onset insulation degradation) \\
C17 & 15 & Flow Rate + Temperature    &              & Irregular bounded oscillation ($\pm$0.8$\sigma$); normal demand variability \\
C18 &  4 & Diff.\ Pressure / Ammeter  &              & Low-amplitude flat signal; stable normal behaviour shared by two sensor types \\
C19 &  4 & Mixed HVAC                 &              & Near-zero flat with single brief negative excursion; likely sensor dropout \\
\bottomrule
\end{tabular}
\end{table*}

Figures~\ref{fig:cluster1}--\ref{fig:cluster4} display the cluster-representative
90-day sensor series for all 20 clusters, and
Tables~\ref{tab:cluster_a}--\ref{tab:cluster_anomaly} summarise the corresponding
measurement categories, cluster sizes, anomaly flags, and time-series pattern signatures.
The following four observations emerge from a joint reading of these figures and tables.

\begin{enumerate}
  \item \textbf{Anomaly pattern diversity.}
        Anomalous clusters exhibit qualitatively distinct fault signatures
        (Figs.~\ref{fig:cluster1}--\ref{fig:cluster4},
         Tables~\ref{tab:cluster_a}--\ref{tab:cluster_anomaly}):
        rapid oscillation with a sustained positive excursion after day~40 in C0
        (Fig.~\ref{fig:cluster1}, Table~\ref{tab:cluster_a});
        progressive upward drift culminating in a sharp spike in C6, and
        high-frequency noise on a rising mean in C7
        (Fig.~\ref{fig:cluster2}, Table~\ref{tab:cluster_b});
        a step-down followed by a sustained low-pressure plateau in C10, and
        a gradually deepening negative deviation in C11
        (Fig.~\ref{fig:cluster3}, Table~\ref{tab:cluster_c});
        and a slow-onset temperature excursion above $+1\sigma$ in C16
        (Fig.~\ref{fig:cluster4}, Table~\ref{tab:cluster_anomaly}).
        This diversity confirms that no single threshold rule covers all fault types;
        a learned model conditioned on text-embedded equipment context is necessary.

  \item \textbf{Normal clusters are physically coherent.}
        Non-anomalous clusters consistently group sensors with shared physical roles
        (Tables~\ref{tab:cluster_a}--\ref{tab:cluster_anomaly}).
        The most striking example is C12 ($n=85$, Ammeter/Diff.\ Pressure;
        Fig.~\ref{fig:cluster3}, Table~\ref{tab:cluster_c}),
        which shows a near-flat trace across 85 distinct $(e,c)$ pairs.
        Other stable clusters---C4 (motor current, Fig.~\ref{fig:cluster1}),
        C9 (operating current, Fig.~\ref{fig:cluster2}),
        C13/C14 (temperature sub-groups, Fig.~\ref{fig:cluster3})---likewise
        exhibit low within-cluster variance, demonstrating that the text embedding
        correctly partitions sensors by physical quantity and operational role.

  \item \textbf{90-day horizon captures slow-onset precursors.}
        Several anomalous signatures become discernible only over multi-week intervals.
        The progressive drift in C6 and C11 (Fig.~\ref{fig:cluster2}, Table~\ref{tab:cluster_b};
        Fig.~\ref{fig:cluster3}, Table~\ref{tab:cluster_c}) and the slow temperature
        excursion in C16 (Fig.~\ref{fig:cluster4}, Table~\ref{tab:cluster_anomaly})
        each require at least 60--90 days of observation to distinguish abnormal from
        normal seasonal variation.
        Short-horizon detection ($\le$30 days) would miss these precursors entirely.
        The 90-day look-ahead label (\texttt{any\_anomaly}) is therefore the primary
        justification for the extended prediction window in v3-0.

  \item \textbf{Frequency cluster (C15) requires equipment-type context.}
        C15 (Inverter Frequency, $n=10$;
        Fig.~\ref{fig:cluster4}, Table~\ref{tab:cluster_anomaly})
        exhibits a characteristic step-function waveform reflecting discrete variable-speed
        motor stages.
        This pattern would be misclassified as abnormal by a threshold-based rule applied
        uniformly across sensor types, yet it is entirely normal for inverter-driven fan
        units operating at fixed speed presets.
        The text embedding $\mathbf{z}$ encodes the equipment-type label and isolates
        them into a dedicated cluster, making the step-function structure
        learnable as a class-conditional normal pattern without feature engineering.
\end{enumerate}

\section{Discussion}
\label{sec:discussion}

\paragraph{Precision--recall trade-off.}
v3-0 prioritises precision (0.992) over recall (0.926) at 30 days.
In a facility management context, false alarms impose direct operational costs (unnecessary
site visits), whereas missed anomalies are typically caught at the 60- or 90-day horizon.
The 60-day model achieves the best recall~(0.963) and F1~(0.967), making it
the recommended operational configuration when the cost of missed anomalies is higher.

\paragraph{Compute, latency, and edge deployment.}
The inference-time pipeline has a deliberately lightweight footprint designed for
\textbf{edge computing deployment} co-located with facility equipment:
\begin{itemize}
  \item \textbf{Statistical features} ($\mathbf{x}$): computed from the 90-day sensor buffer.
        CPU-only; $<$1~ms per sample.
  \item \textbf{Text embedding} ($\mathbf{z}$): a single dictionary lookup into the
        pre-cached \texttt{.npz} ($\approx$0.1~ms). No GPU or language model required at runtime.
  \item \textbf{TTM embedding} ($\mathbf{y}$): optional at inference if the cached
        \texttt{test\_ttm\_embeddings.npz} is available; otherwise $<$5~ms on GPU
        or $<$80~ms on CPU.
  \item \textbf{LightGBM inference}: serialised model file $<$3~MB;
        $<$1~ms per sample on any CPU.
\end{itemize}
The total inference latency is \textbf{under 2~ms} on CPU-only hardware when embedding caches
are pre-built, making the pipeline deployable on NVIDIA Jetson Orin, Intel NUC,
or equivalent edge servers embedded in building management systems.
Cloud connectivity is required only for the initial one-time offline embedding step;
all subsequent anomaly scoring runs fully air-gapped.

\paragraph{FAISS Index as a Data-Governance Tool.}
A persistent challenge in industrial facility management is \emph{entity name heterogeneity}:
the same physical equipment or measurement item is recorded under different names across
work-order systems, sensor databases, and master tables (\textit{nayose} problem).
Traditional resolution relies on manual curation, which is costly and error-prone at scale.

The FAISS IVFFlat index built over the 1,024-dimensional multilingual-e5-large embeddings
of all 580 $(e,c)$ pairs offers a data-driven alternative.
Because semantically equivalent descriptions---even when expressed with different abbreviations,
vendor codes, or languages---are mapped to nearby points in embedding space,
$k$-nearest-neighbour retrieval can surface candidate matches for human review or
automated deduplication:
\begin{itemize}
  \item \textbf{Cross-system entity linking}: given a new equipment entry from an external
        work-order system, a top-$k$ query identifies the closest existing master records,
        enabling semi-automated \textit{nayose} without hand-written matching rules.
  \item \textbf{Anomaly pattern transfer}: once entities are linked,
        anomaly signatures from well-monitored equipment can be propagated to
        newly registered or data-sparse units via neighbour-weighted feature augmentation.
  \item \textbf{Master data quality monitoring}: low-similarity nearest-neighbour distances
        serve as an outlier score for entries that may be duplicates, typos, or
        mis-categorised records---providing a continuous data-quality signal.
\end{itemize}
This positions the FAISS index not merely as a retrieval accelerator but as a
\textbf{semantic anchor for equipment data governance}, complementing the prediction pipeline
with a long-term asset for master data management.

\paragraph{Generalisation to other facility types.}
This study targets HVAC air-conditioning equipment (pumps, chillers, air-handling units,
distributed AC, and mechanical systems) at a single facility.
The cluster structure in Figure~\ref{fig:umap} demonstrates that free-form Japanese
check-item text alone is sufficient to form semantically coherent groups,
suggesting the approach is not inherently HVAC-specific.
However, extending the system to other facility domains---manufacturing lines,
data-centre cooling, or building electrical systems---introduces new challenges:
the anomaly label distribution may differ substantially,
the statistical feature set requires domain-specific validation,
and the TTM embeddings must be fine-tuned or replaced if sensor sampling rates
or physical units change.
A systematic transfer-learning study across facility types is left for future work.

\paragraph{Open-Source Stack and Reproducibility.}
The entire pipeline is built exclusively on open-source components
(Table~\ref{tab:oss}), enabling royalty-free deployment in commercial facility management systems.
All source code is released at
\url{https://github.com/tk-yasuno/stat_tsfm_text_fusion_gbdt}
under the Apache~2.0 License.

\begin{table*}[t]
\centering
\caption{Full open-source stack. All components are freely usable in commercial settings.}
\label{tab:oss}
\footnotesize
\begin{tabular}{llll}
\toprule
Component & Model / Library & License & Role \\
\midrule
TS encoder & IBM Granite TTM~\cite{ekambaram2024ttms} & Apache~2.0 & $\mathbf{y}$ \\
LoRA adapter & PEFT (HuggingFace) & Apache~2.0 & Fine-tuning \\
Text encoder & multilingual-e5-large~\cite{wang2024multilingual} & MIT & $\mathbf{z}$ \\
Classifier & LightGBM~\cite{ke2017lightgbm} & MIT & Prediction \\
\bottomrule
\end{tabular}
\end{table*}

\section{Conclusion}
\label{sec:conclusion}

We presented a Triplet Feature Fusion framework for equipment anomaly prediction that
combines statistical features, a LoRA-adapted time-series Transformer, and multilingual text
embeddings of Japanese equipment master records.
On a 64-equipment HVAC dataset, the triplet model achieves 0.992 precision and 0.998 ROC-AUC
at the 30-day horizon, reducing false positives by 83\% compared to a strong dual-feature baseline.

Cluster analysis of 278 unique text embeddings identifies 20 semantically coherent equipment
groups with distinct fault archetypes, confirming that the text vector provides a genuine
equipment-type prior rather than noise.
Design lessons derived from the full v1.0--v3.0 development trajectory highlight the importance
of statistical features as primary discriminators, the complementary role of Transformer
embeddings for non-linear temporal patterns, and the unique value of text conditioning for
false positive suppression.

Future directions include:
(i) integration of domain-specific statistical knowledge---such as hydrological inflow
regression~\cite{yasuno2018dam} and ensemble-weighted flood forecasting~\cite{yasuno2022floodinflow}---as
a template for encoding physics-informed feature priors into the triplet framework,
enabling richer anomaly signals for specialised facility domains;
(ii) extension to multi-facility datasets for cross-domain generalisation;
(iii) \textbf{edge-native deployment}---packaging the inference pipeline as an OCI container
for NVIDIA Jetson or ARM-based building management controllers,
enabling real-time anomaly scoring directly at the sensor gateway
without cloud round-trips;
and (iv) \textbf{fourth-modality feature extension}---augmenting the triplet
$\mathbf{h} = [\mathbf{x};\mathbf{y};\mathbf{z}]$ with an additional feature branch
$\mathbf{w}$ encoding complementary evidence sources.
On the \emph{static} side, $\mathbf{w}$ could incorporate expert visual-inspection remarks
encoded as text embeddings, or discrete damage-history flags derived from earthquake and
flood disaster records~\cite{yasuno2018dam,yasuno2022floodinflow}.
On the \emph{dynamic} side, $\mathbf{w}$ could carry acoustic time-series embeddings
that surface abnormal-sound signals invisible to numeric sensors,
or CNN-extracted representations of inspection images targeting critical mechanical
components---enabling a richer quadruplet fusion $[\mathbf{x};\mathbf{y};\mathbf{z};\mathbf{w}]$
for fault types that manifest primarily in non-numeric modalities.

\section*{Acknowledgements}

The author thanks colleagues at YACHIYO Solutions Co., Ltd.\ for discussions on
facility management practices.
All experiments were conducted on a personal workstation outside working hours.
All sensor records and equipment master data are fully anonymised internal operational data
handled in accordance with the company's data-governance policy;
no confidential or proprietary data were used.

\section*{Source Code Availability}

The complete implementation including training scripts and experimental
configuration files is available as open source at:
\url{https://github.com/tk-yasuno/stat_tsfm_text_fusion_gbdt}

\bibliographystyle{unsrt}
\bibliography{triplet_feature_anomaly_2026}

\end{document}